\documentclass[sn-basic]{sn-jnl}

\jyear{2021}%

\theoremstyle{thmstyleone}%
\theoremstyle{thmstyletwo}%

\theoremstyle{thmstylethree}%

\usepackage{booktabs}
\usepackage{soul}
\usepackage{amsmath}
\usepackage{amssymb}
\usepackage{makecell}
\usepackage{bbding}
\usepackage{pifont}
\usepackage{bigstrut}
\usepackage{color}
\usepackage{algorithm}
\usepackage{algpseudocode}
\usepackage{multirow}

\begin{document}

\title[Deep Multimodal Fusion for Generalizable Person Re-identification]{Deep Multimodal Fusion for Generalizable Person Re-identification}


\author*[]{\fnm{Suncheng} \sur{Xiang}}\email{xiangsuncheng17@sjtu.edu.cn}

\author[]{\fnm{Hao} \sur{Chen}}\email{958577057@sjtu.edu.cn}

\author[]{\fnm{Wei} \sur{Ran}}\email{rw1997@sjtu.edu.cn}

\author[]{\fnm{Zefang} \sur{Yu}}\email{yuzefang@sjtu.edu.cn}

\author[]{\fnm{Ting} \sur{Liu}}\email{louisa\_liu@sjtu.edu.cn}

\author[]{\fnm{Dahong} \sur{Qian}}\email{dahong.qian@sjtu.edu.cn}

\author[]{\fnm{Yuzhuo} \sur{Fu}}\email{yzfu@sjtu.edu.cn}

\affil[]{\orgname{Shanghai Jiao Tong University}, \orgaddress{\city{Shanghai}, \postcode{200240}, \country{China}}}



%


\abstract{Person re-identification plays a significant role in realistic scenarios due to its various applications in public security and video surveillance. Recently, leveraging the supervised or semi-unsupervised learning paradigms, which benefits from the large-scale datasets and strong computing performance, has achieved a competitive performance on a specific target domain. However, when Re-ID models are directly deployed in a new domain without target samples, they always suffer from considerable performance degradation and poor domain generalization. To address this challenge, we propose a Deep Multimodal Fusion network to elaborate rich semantic knowledge for assisting in representation learning during the pre-training.
Importantly, a multimodal fusion strategy is introduced to translate the features of different modalities into the common space, which can significantly boost generalization capability of Re-ID model. As for the fine-tuning stage, a realistic dataset is adopted to fine-tune the pre-trained model for better distribution alignment with real-world data. Comprehensive experiments on benchmarks demonstrate that our method can significantly outperform previous domain generalization or meta-learning methods with a clear margin.
Our source code will also be publicly available at \url{https://github.com/JeremyXSC/DMF}.}




\maketitle

\section{Introduction}
\label{sec1}
Person re-identification aims to match a specific person in a large gallery with different cameras and locations, which has attracted extensive research attention thanks to its practical importance in the surveillance system. With the development of deep convolution neural networks, person Re-ID methods have achieved remarkable performance in a supervised manner. However, these supervised approaches are hardly applicable in practice due to expensive labeling costs and also suffer from severe performance degradation on an unseen target domain, which is mainly caused by the domain gap between different backgrounds, camera angles, and camera styles, is now becoming the main challenges for Re-ID community. To solve this problem, previous research works mainly overcome this difficulty from the perspective of unsupervised domain adaptation with prior knowledge from target domain, which either eliminate the data distribution discrepancy across source and target domain with generative adversarial networks~\citep{wei2018person,deng2018image,xiang2020unsupervised}, or deploys advanced clustering algorithm to generate pseudo-labels for unsupervised target images during training period~\citep{wang2022uncertainty,xiang2022learning}. Although these approaches can achieve promising progress, their performance deeply relies on the learned prior knowledge from the target domain. Consequently, when Re-ID models are directly deployed in new scenarios, they always suffer from considerable performance drop and poor domain generalization.
In essence, the performance of generalizable person Re-ID suffers from considerable
degradation when compared with the supervised or semi-supervised setting, which is mainly resulted from the variation of illumination, background, camera angle, resolution, \textit{etc.} between source domain and target domain. So how we design training strategy to be well applicable for domain generalization, has to be handled carefully.

\begin{figure}[!t]
\centerline{\includegraphics[width=1.0\linewidth]{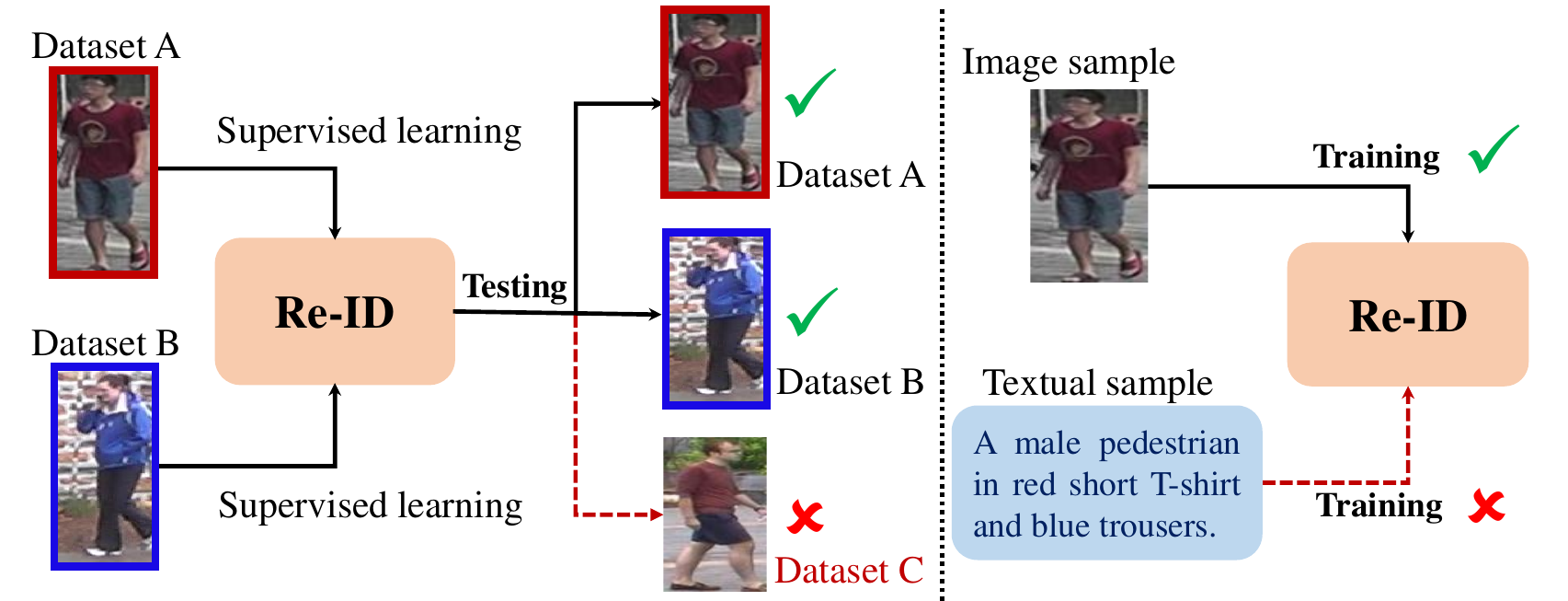}}
\caption{The drawbacks of previous Re-ID approaches: (1) traditional Re-ID approaches are ineffective on Dataset C due to the difference between cameras and distribution bias among different domains (\textbf{Left part}); (2) traditional Re-ID model fails to explore the potential of the textual annotations during training (\textbf{Right part}). In this work, we argue that these two characters play an important role for training a discriminative Re-ID model.}
\label{fig1}
\end{figure}

The recently emerged research of Vision Transformers~\citep{han2022survey,he2021transreid,dosovitskiy2020image,pei2022transformer} have gained increasing attention from the public due to its promising performance in image match tasks, which adopts the pure transformer to learn robust feature representation on image-level for Re-ID research. Unfortunately, previous Re-ID approaches expose some notable shortcomings on generalizable Re-ID events. As shown in Figure~\ref{fig1}, these works mainly focus on the image samples to assist the pre-training or fine-tuning of Re-ID model, which fails to explore the potential of the multimodal dataset (\textit{e.g.} Image data \& Textual annotations) for Re-ID training.
To relieve this dilemma, many researchers have laid emphasis on constructing texture data which describes the attributes of pedestrians in text modality, for example, \cite{xiang2021less} manually construct a large-scale person dataset named FineGPR with fine-grained textual descriptions of pedestrian appearances. Even though these early attempts can promote the development of deep multimodal learning in Re-ID community, the generalization capability of Transformers with multimodal datasets for image matching is still unknown.

To address these challenges, in this paper, we propose \textbf{DMF}, a \textbf{D}eep \textbf{M}ultimodal \textbf{F}usion network that incorporates fine-grained and high-level semantic textual
features during training. To extract high-level semantic features of pedestrians, we adopt a large-scale synthetic dataset named FineGPR with fine-grained attribute annotations for model's pre-training, which has rich texture annotations including 14 attributes describing pedestrian identities and 4 attributes describing environment information. Moreover, to further enhance the generalization ability of deep model, we intentionally employ a standard multimodal transformer encoder to translate the features of image and text into the common space for feature fusion. Specifically, the fused space is an expanded space with two modalities feature, which can promote the robust learning of our Re-ID model. Then a vanilla transformer is adopted as the backbone network to extract the features in the expanded feature space to calculate cross-attention. Finally, our pre-trained encoders are fine-tuned with real-world datasets for distribution alignment on down-stream Re-ID task.

To this end, our proposed method can not only learn fine-grained information from different modalities; but also effectively reduce the style
differences between different domains.
After the pre-training, we fine-tune the pre-trained model with image samples from realistic dataset, so that the weight of feature extractor can align with the distribution of real-world closely. To the best of our knowledge, we are the first to introduce multimodal features to vision transformer on generalizable person Re-ID event. With this approach, we proved that a vanilla  transformer can achieve competitive performance with multimodal dataset on generalizable image retrieval.
Compared with existing vision transformer based method~\citep{liao2021transmatcher,he2021transreid}, our DMF is different from them in terms of two perspective: dataset input and model structure: (1) our work adopts multimodal data to learn fine-grained information from different modalities, while previous vision transformer based methods only employ single-modal data (\textit{e.g.} image feature) to train deep model; (2) traditional Transreid~\citep{he2021transreid} and TransMatcher~\citep{liao2021transmatcher} method directly inherit the merits of traditional vision transformer on downstream task, which fails to explore the potential of multimodal datasets on vision transformer. In contrast, our DMF network deploys a pre-trained encoder to translate the data of the two modalities into common feature space, and introduces abundant semantic knowledge to assist in the pre-training of transformer backbone, which can significantly eliminate the gap between different domains.

In this work, our major contribution can be summarized into three points:
\begin{itemize}
		\item[$\bullet$] We propose a deep multimodal fusion method named \textbf{DMF}, which elaborates rich semantic knowledge to assist in training for generalizable person Re-ID task.
		\item[$\bullet$] A simple but effective multimodal fusion strategy is introduced to translate data of different modality into the common feature space, which can significantly eliminate the gap between different domains.
		\item[$\bullet$] Comprehensive experiments on benchmarks demonstrate that our proposed method can achieve state-of-the-art performance on generalizable person Re-ID events.
\end{itemize}

In the rest of the paper. we first review some related works of person re-ID datasets and methods in Section~\ref{sec2}. Then in Section~\ref{sec3}, we give the learning procedure of the proposed DMF method and feature fusion strategy. Extensive evaluations compared with state-of-the-art methods and comprehensive analyses of the proposed approach are reported in Section~\ref{sec4}. Conclusion and future work are given in Section~\ref{sec5}.

\section{Related Works}
\label{sec2}

\subsection{Person Re-ID datasets}
Many researchers from both academy and industry have made considerable contributions to build high-quality Re-ID datasets. Among them, \cite{zhang2021unrealperson} built the UnrealPerson pipeline that can make full use of synthesized image data towards
a powerful Re-ID algorithm. \cite{sun2019dissecting} introduce a large-scale synthetic data engine PersonX that can generate images under controllable cameras and environments.
\cite{wang2020surpassing} collect a virtual dataset named RandPerson with 3D characters which contains 1,801,816 synthetic images of 8,000 identities. However, these datasets are either in a small scale or lack of diversity, few of them provide rich attribute annotations, which cannot satisfy the need for multimodal learning in person Re-ID task. To relieve this dilemma, \cite{xiang2021less} manually construct a large-scale synthetic dataset named FineGPR with fine-gained attributes annotations, which is also affiliated with a diversified FineGPR-C caption dataset~\citep{xiang2021rethinking} for semantic-based pre-training. Compared with previous realistic datasets, FineGPR has a more evenly distribution of attributes, which is closer to the real world. Despite this, there is currently no suitable research to adopt the accurately semantic annotations of FineGPR dataset~\citep{xiang2021less} to reduce the domain gap between different domains and enhance the generalization ability of deep model. So, further progress related to multimodal deep learning on Re-ID event is needed.

\subsection{Person Re-ID methods}

For adaptive person re-identification task, a na\"{\i}ve approach is to label samples in target domain with pseudo-labels on the basis of DBSCAN~\citep{ester1996density} clustering algorithm, and then the generated pseudo-labels are used to guide the learning of image representations. It is worth mentioning that unsupervised learning methods do not need to know the labels of the samples in the target domain, but the overall distribution or prior knowledge of the target domain in advance, which is a prerequisite for clustering. However, for an irrelevant domain without any information, the prediction performance of these models is far from satisfactory.

Recently, benefited from the advantages of transformer~\citep{vaswani2017attention,jia2022learning,zhong2020self,vaswani2017attention}, the performance of person re-identification in supervised learning has been greatly improved. Relying on this, the person Re-ID model obtains more robust feature extraction and more reliable metric learning. For example, \cite{dosovitskiy2020image} have shown that a pure transformer ViT applied directly to sequences of image patches can perform very well on image classification tasks. To advance the research of Re-ID task with the merits of transformer network, \cite{he2021transreid} propose a pure transformer-based object Re-ID framework named TransReID, which uses camera information to improve the performance of person re-identification into a new level. \cite{baevski2022data2vec} introduce a general framework data2vec for self-supervised learning in speech, vision and language. Instead of predicting modality-specific objects, data2vec predicts contextualized latent representations that contain information from the entire input. Additionally, \cite{kim2021vilt} present a Vision-and-Language Transformer (ViLT) to focus more on the modality interactions inside the transformer module. \cite{liao2021transmatcher} introduce query-gallery cross-attention in the vanilla Transformer to obtain notable performance improvements. However,
common vision transformer based methods pay litter attention to the interactions among the data in different modalities. Importantly,
all these methods neither take the multimodal information into account for generalizable Re-ID task, nor learn any texture information for mining the gap between different domain in advance, which undoubtedly degradates the learning ability of Transformers for generalizable image matching.
To address this problem, in this work, we take a big step forward and propose a deep multimodal fusion method to introduce rich semantic knowledge to assist in training of transformer backbone, and then we fine-tune the pre-trained encoders with real-world datasets for more robust feature representation.

\begin{figure*}[!t]
\centerline{\includegraphics[width=1.0\linewidth]{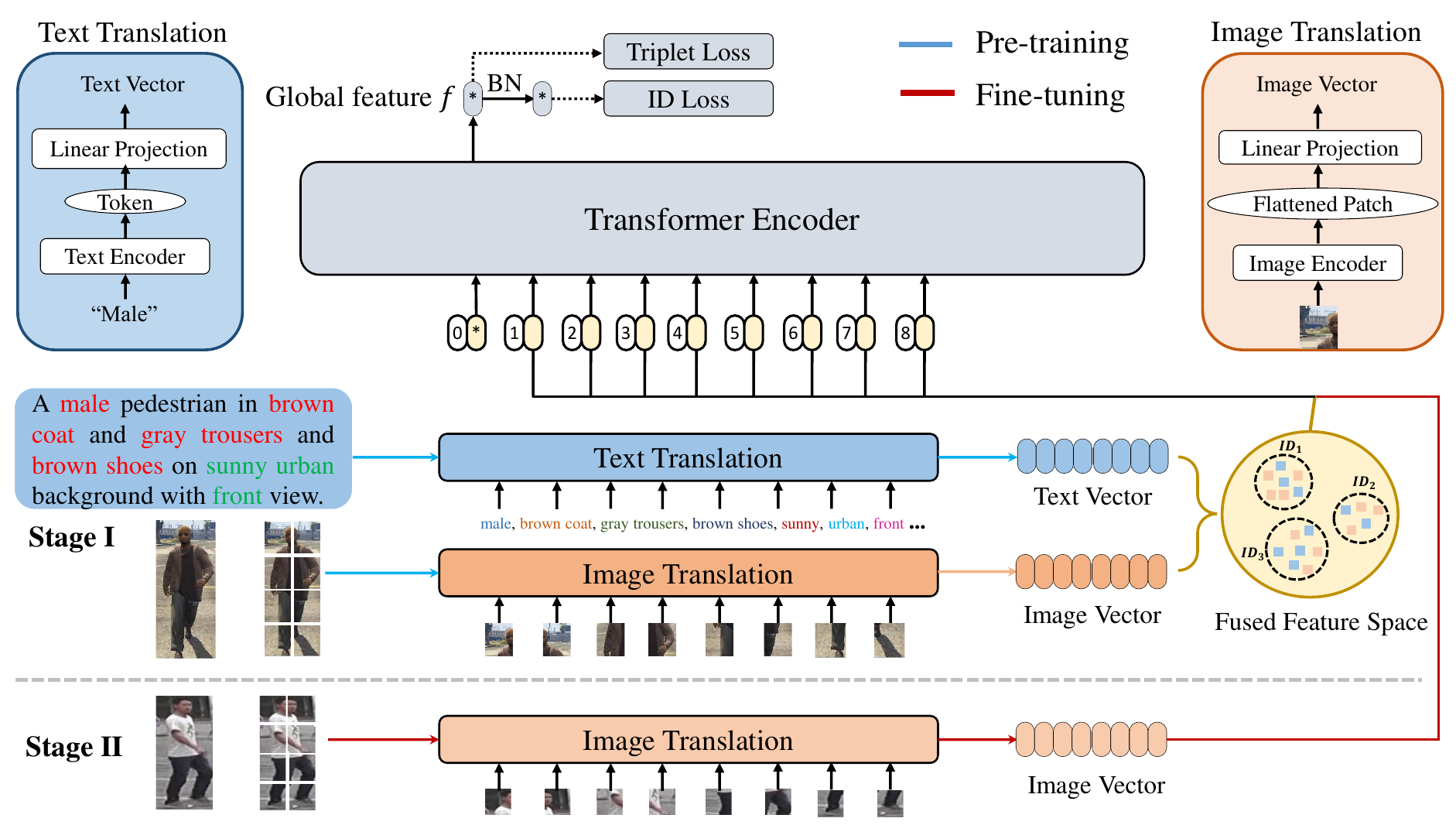}}
\caption{The Pre-training-and-fine-tuning process of \textbf{DMF} network. The backbone is based on Vision Transformer (ViT). The text translation and image translation, which translate the textual annotations and images to the same feature space, are based on pre-trained encoder and linear projection. After transformer layers, global feature is extracted and introduced to calculate triplet loss, while feature of another branch after BNNeck (BN) is introduced to calculate ID loss. Zoom in for the best view.}
\label{fig2}
\end{figure*}



\section{Methodology}
\label{sec3}
In this section, we firstly give the problem definition of generalizable person Re-ID task. Then we introduce the deep multimodal fusion network (DMF), as illustrated in Figure~\ref{fig2}. Finally, we elaborate more details of multimodal fusion strategy.

\subsection{Problem Definition}

We begin with a formal description of generalizable person Re-ID problem. Assuming that we are given with $K$ source domains $\mathcal{D}=\left\{\mathcal{D}_k\right\}_{k=1}^K$. Each source domain contains its own label space or person identities $\mathcal{D}_k=\left\{\left(\boldsymbol{x}_i^k, y_i^k\right)\right\}_{i=1}^{N_k}$, where $N_{k}$ is the number of images in the source domain $\mathcal{D}_k$. Each sample $\boldsymbol{x}_i^k \in \mathcal{X}_k$ is associated with an identity label $y_i^k \in \mathcal{Y}_k=\left\{1,2, \ldots, M_k\right\}$, where $M_{k}$ is the number of identities in the source domain $D_{k}$.
During the training stage, we train a domain generalization model using the aggregated image-label pairs of all source domains.
In the testing stage, we perform a retrieval task on unseen target domains without additional model updating.


\subsection{Deep Multimodal Fusion Network}

Intuitively, there exists a serious domain gap of image data in different domains. However, multimodal data from different domain may not have style differences if the pedestrians in different domains are described by another modality, \textit{e.g.}, text. Aim at bridging the gap between different domain and improve the performance of generalizable Re-ID event, we propose a deep multimodal fusion network to introduce rich semantic knowledge to assist in Re-ID training.
To be more specific, we split the whole DMF network into two stages: \textbf{pre-training} and \textbf{fine-tuning}, and the overall framework of our method is illustrated in Figure~\ref{fig2}.

Since each image from FineGPR dataset corresponds to a textual description. As for the \textbf{pre-training stage}, we fuse the samples of the two different spaces to obtain an expanded feature space, which contains the feature from both image and text data. Our goal is to pre-process the data of the two modalities as tokens and flattened patches respectively, then translate them through a pre-trained encoder~\citep{baevski2022data2vec}, which can equally translate data of different modalities without learning their modalities. Finally, the samples of different modalities are translated to the same subspace through linear projection layer with same dimension. More training details related to our proposed DMF method are depicted in Algorithm~\ref{alg1}.

\begin{algorithm}[!t]
	\renewcommand{\algorithmicrequire}{\textbf{Input:}}
	\renewcommand{\algorithmicensure}{\textbf{Output:}}
	\caption{The proposed DMF method}
	\label{alg1}
	\begin{algorithmic}[1]
		\Require
        Image vector  $S_i$ for pre-training, texture vector  $S_t$ for pre-training, image vector  $S_f$ for fine-tuning, iteration $n_1$ for pre-training, iteration $n_2$ for fine-tuning;
		\Ensure
		Multi-modal Re-ID model $f(x)$;
		\State \textbf{stage I: Pre-training}
		\For{$iter\leq n_1$}
		\State Randomly select a fusion sample $x \in S_i\cup S_t $
		\If{$x \in S_i$}
		\State Do image translation $x \rightarrow v_i$
		\State $v_i \leftarrow v_i.append(v_{cls})+p_i$
		\State Optimize $f$ with $v_i$
		\Else
		\State Do text translation  $x \rightarrow v_t$
		\State $v_t \leftarrow v_t.append(t_{cls})+p_t$
		\State Optimize $f$ with $v_t$
		\EndIf
		\EndFor
		\State \textbf{stage II: Fine-tuning}
		\For{$iter\leq n_2$}
		\State Randomly select an image sample $x \in S_f $
		\State Do image translation $x \rightarrow v_i$
		\State $v_i \leftarrow v_i.append(v_{cls})+p_i$
		\State Optimize $f$ with $v_i$
		\EndFor
	\end{algorithmic}
\end{algorithm}

Assuming that the dataset for fine-tuning is $S_f$, during the \textbf{fine-tuning stage}, we firstly select a sample $x\in S_f$ and split the images into $N$ parts.
Then we input the image patches to the pre-trained image encoder and flatten them to low-dimensional vector by a linear projection layer (see Eq.~\ref{eq3}). Consequently, the low-dimensional vector $v_j$ is mapped to $D$-dimensional $F(v_j) \in R^{1\times {D}}$ through the pre-trained linear projection layer.
Finally, an extra learnable [cls] embedding token denoted as $x_{cls}$ is prepended to the input sequences, and spatial information is incorporated by adding learnable position embedding $p_{i}$, which are fed to transformer encoder all together. It is worth mentioning that the weight of the image translation module during fine-tuning is shared with pre-training stage. During the fine-tuning stage, we only adopt datasets different from the target domain for representation learning, which can effectively make the model pay more attention to the general distribution of dataset that is closer to the real scenarios$\footnote[1]{We only use visural images to fine-tune the pretrined Re-ID model since the realistic datasets (\textit{e.g.} Market-1501) have no textual caption.}$.



In essence, many previous works~\citep{xiang2022learning,luo2019strong} have been found that
performing training with multiple losses has great potential to learn a robust and generalizable Re-ID model.
Motivated by this, we adopt identification loss~\citep{zheng2017discriminatively} and triplet loss~\citep{hermans2017defense} as basic constraints based on the Transformer backbone, which can translate the images and text to the same feature space for reliable feature embedding. For a triplet couple \{$x_a$, $x_p$, $x_n$\}, the soft-marginal triplet loss $\mathcal{L}_{triplet}$ can be calculated as:

\begin{equation}\label{eq1}
\mathcal{L}_{Triplet}=\log \left[1+\exp \left(\left\|x_a-x_p\right\|_2^2-\left\|x_a-x_n\right\|_2^2\right)\right]
\end{equation}
where $x_a$, $x_p$ and $x_n$ denote the anchor image, positive sample and negative sample respectively.

In this paper, we propose to jointly learn deep multimodal feature using ID loss and triplet loss in a training batch. To be more specific, there are an image sample set $S_i$ and texture sample $S_t$ for pre-training, image sample set $S_f$ for fine-tuning. Finally, the overall objective loss function in a training batch is expressed as:

\begin{equation}
\mathcal{L}_{total}= \mathcal{L}_{ID} + \mathcal{L}_{Triplet}
\label{eq2}
\end{equation}

\subsection{Multimodal Feature Fusion}
\begin{figure}[!t]
\centerline{\includegraphics[width=1.0\linewidth]{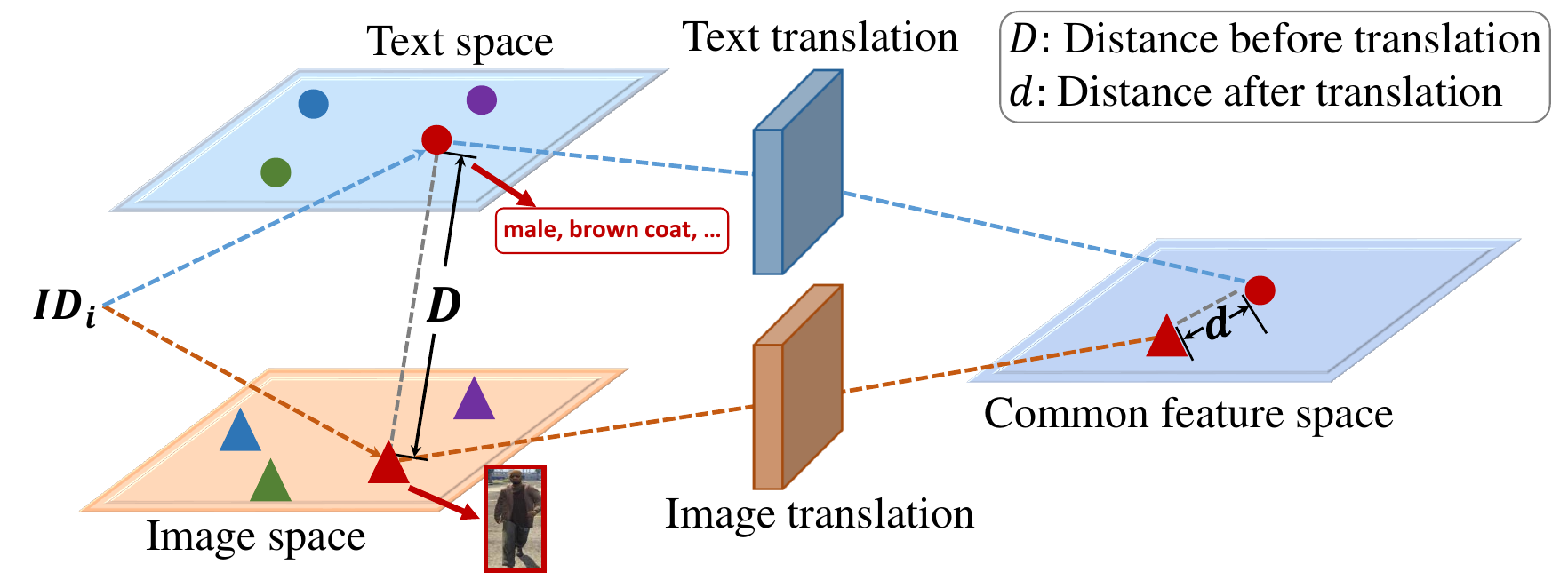}}
\caption{Our multimodal fusion strategy of DMF network. The initial samples are distributed in image space and text space respectively, and the distance between textual sample and image sample from the same identity is represented as $D$ in original feature space. After translation, $D$ is replaced by the distance $d$ in common space. Obviously, distance $d$ among same identity is much closer than $D$.}
\label{fig3}
\end{figure}

In order to translate the image feature to the fused feature space, we pre-process the image samples before inputting them to the Transformer network. As shown in Figure~\ref{fig3}, vector-form samples from the same identity are pulled closer in the same feature space after the feature translation from different modalities. In particular,
for an image sample $x\in R^{H\times W\times C}$, we first split the image into $N$ parts, where $H$, $W$, $C$ denote height, width and number of channels respectively. Each part of the image has the same value in terms of width and height aspects, by which we can get $N$ patches of the image
$\{x_j\in R^{H'\times{W'}\times{C}}, j=1, 2, ..., N\}$, where $H'$ and $W'$ denote the height and width of the image patches. After image pre-processing, we input the image patches to the pre-trained image encoder~\citep{baevski2022data2vec}, which can effectively flatten image features to low-dimensional vectors.
The pre-trained  image encoder can be expressed as:
\begin{equation}
v_j=Image2vec \left(x_j \right), j=1,2,...,N
\label{eq3}
\end{equation}
where $v_j\in R^{1\times {D_i}}$ and $x_j$ denotes the $j^{th}$ part of the patches.
Based on the output of pre-trained image encoder, we continuously design a linear projection layer to map the low-dimensional vector $v_j$ to $D$-dimensional $F(v_j) \in R^{1\times {D}}$, and an extra learnable $\left[cls \right]$ embedding token denoted as $v_{cls}$ is also prepended to the input sequences: $[v_{cls}, F(v_1), F(v_2),..., F(v_N)]$, which gives rise to the vector $v_i$:
\begin{equation}
v_i = [v_{cls}, F(v_1),F(v_2),...F(v_N)]+p_i
\label{eq4}
\end{equation}
where  $p_i\in R^{(N+1)\times D}$ represents the position embedding which encodes positional information of image samples, $F(\cdot)$ denotes the linear projection which can translate the tokens from $D_i$-dimensional space to $D$-dimensional space.
Finally,  the vector $v_i$ is fed to the transformer encoder for model pre-training.

Similarly, given a textual description $t$, firstly we extract the key words of fine-grained attribute annotations, and then introduce them to the pre-trained text encoder~\citep{baevski2022data2vec} (see Eq.~\ref{eq3}), after that we obtain \textit{N} tokens word by word  $\{t_k\in R^{1\times{D_t}}, k=1,2,...,N\}$. These tokens are $D_t$-dimensional vectors in textual space, which cannot be directly used as feature embedding. To relieve this dilemma, we design a linear projection layer to transform the dimension of text space to $D$-dimensional space $R^{1\times {D}}$, which matches the image feature space exactly. These tokens after the linear projection are spliced to a new vector denoted as $[L(t_1), L(t_2),..., L(t_N)]$, then it is fed to the transformer encoder to predict the identity of the original text. More importantly, an extra learnable $\left[cls \right]$ embedding token named as $t_{cls}$ is prepended to the text vector, and a learnable position embedding $p_{t}$ is added to each vector to represent the correlation between the positions of different tokens. Consequently, the text vector fed into transformer encoder can be described as:
\begin{equation}
v_t = [t_{cls}, L(t_1), L(t_2),..., L(t_N)]+p_t
\label{eq5}
\end{equation}
where $p_t \in R^{(N+1)\times D}$denotes the position embedding which encodes positional information of text samples, $L(\cdot)$ denotes the linear projection which can translate the tokens from $D_t$-dimensional space to $D$-dimensional space.

Theoretically speaking, our DMF method can obtain better performance with multimodal datasets than training on single-modal datasets.
Assuming that the feature space of image vector after linear projection is $S_i$, as for $\forall v_i \in S_i$,  $v_i \in R^{(N+1)\times D}$, therefore $S_i \subset R^{(N+1)\times D}$. Similarly, the space of text vector $S_t \subset R^{(N+1)\times D}$. Both of them are subsets of the
$(N+1)\times D$-dimensional space. In other words, we extend our original feature space from $S_i$ to $S_i \cup S_t$, and the samples in the extended space are all labeled, which can effectively guide the Re-ID model to learn more external information.
In order to fuse the two feature spaces more effectively, we set locally position embedding and class embedding for each space, and a modality-specific linear projection is introduced to flexibly translate the vector to the same feature space. To be more specific, the samples of images and text are independent in the terms of the fused feature space, so that the model can learn domain invariant and generalizable feature during multimodal training. Consequently, our proposed method can greatly enhance the generalization ability of Re-ID model and achieve the competitive performance in generalizable Re-ID event.

Overall, through the cross-entropy loss and triplet loss via multimodal fusion, the discriminative ability of deep multimodal fusion network is enhanced in the domain generalization process, and more reliable or hard samples are involved during training, eventually leading to the performance improvement of the deep multimodal fusion network.

\section{Experiments}
\label{sec4}

\subsection{Datasets}
\label{sec4.1}
We use one synthetic dataset FineGPR~\citep{xiang2021less} to peforrm pre-training and adopt four public datasets, \textit{i.e.}, Market-1501~\citep{zheng2015scalable}, DukeMTMC-reID~\citep{ristani2016performance}, MSMT17~\citep{wei2018person} and Randperson~\citep{wang2020surpassing} to perform fine-tuning. We test our model on Market-1501, CUHK03~\citep{li2014deepreid}, MSMT17 respectively.

\textbf{Market-1501} contains 32,668 labeled images of 1,501 identities. There are 6 cameras from campus in Tsinghua University. As official setting, 751 identities are adopted as trainset and the rest 750 identities are used for testing. The query contains 3,368 images.

\textbf{DukeMTMC-reID} has 36,411 labeled images of 1,404 identities, which is collected in the winter of Duke University from 8 different cameras, which contains 16,522 images of 702 identities for training, and the remaining images of 702 identities used for testing, including 2,228 images as query and 17,661 images as gallery, the bounding-box of this dataset are all manually annotated.

\textbf{CUHK03} contains 14,097 images of 1,467 identities, which are collected in the Chinese University of Hong Kong with only 2 cameras. This dataset has two settings of labelling: human labeled bounding boxes and DPM detected bounding boxes. In this paper, we mainly use CUHK03 (detected) in our experiment because it is much closer to practical scenarios.

\textbf{MSMT17} contains 126,441 images of 4,101 identities, which includes 32,621 training images from 1,041 identities. For the testing set, 11,659 bounding boxes are used as query images and other 82,161 bounding boxes are used as gallery images.

\textbf{FineGPR} is a large-scale synthetic dataset with fine-grained attribute annotations, which contains 2,028,600 synthesized person images of 1,150 identities with 14 foreground attributes and 4 background attributes. We randomly select 11 images from each identity to form our pre-training dataset with a total of 12,650 images and texts.

\textbf{RandPerson} contains 1,801,816 synthesized person images of 8,000 identities. On the basis of Unity3D engine, this dataset has synthesized person model which includes different viewpoints, poses, illuminations, backgrounds, resolutions and textures. However, this dataset does not contain the fine-grained attributes so it cannot satisfy the need of multimodal pre-training.

\textbf{Evaluation Protocols} For evaluation metric, we follow the conventions in the
Re-ID community and adopt mean Average Precision (mAP) and Cumulative Matching Characteristics (CMC) at Rank-1 for evaluation on Re-ID task.

\subsection{Experiment settings}
In this paper, we follow the standard training procedure in Transreid~\citep{he2021transreid}, and all person images are resized to 256 $\times$ 128 for image processing. The batch size of input samples is set to 64. As for triplet selection, we randomly selected 16 persons and sampled 4
images for each identity in Eq.~\ref{eq1}. SGD optimizer is also employed with a momentum of 0.9 and the weight decay is set as 1e-4. The learning rate is initialized as 0.01 with cosine learning rate decay.
Additionally, we set maximum iteration rounds $n_{1}=120$ and $n_{n}=120$ until it reaches convergence state. And all experiments are conducted on a server equipped with one RTX 3090 GPU and a Intel Xeon Gold 6240 CPU.

\begin{table}[!t]
  \centering
  \caption{Ablation study of different pre-training settings (\textit{e.g.} \textbf{DMF w/o Image}, \textbf{DMF w/o Text} and \textbf{DMF}) from FineGPR dataset. Note that the pre-trained model is then fine-tuned on DukeMTMC-reID dataset for downstream Re-ID task.}
  \small
  \setlength{\tabcolsep}{0.87mm}{
    \begin{tabular}{lcccccccc}
    \toprule
    \multirow{2}[4]{*}{Pre-training} & \multirow{2}[4]{*}{Text data} & \multirow{2}[4]{*}{Image data} & \multicolumn{2}{c}{Market-1501$\uparrow$} & \multicolumn{2}{c}{CUHK03$\uparrow$} & \multicolumn{2}{c}{MSMT17$\uparrow$} \\
\cmidrule{4-9}          &       &       & Rank-1 & mAP   & Rank-1 & mAP   & Rank-1 & mAP \\
    \midrule
    Base. & $\times$     & $\times$     & 54.2  & 30.1  & 11.4  & 10.4  & 9.1   & 8.7  \\
    DMF   & $\checkmark$     & $\times$     & 64.1  & 38.7  & 14.3  & 13.1  & 13.7  & 12.5  \\
    DMF   & $\times$     & $\checkmark$     & 74.9  & 46.3  & 19.3  & 17.9  & 17.5  & 16.0  \\
    DMF   & $\checkmark$     & $\checkmark$     & 80.2  & 52.6  & 21.2  & 19.4  & 20.8  & 19.3  \\
    \bottomrule
    \end{tabular}}%
  \label{table1}%
\end{table}%

\begin{figure*}[!t]
\centerline{\includegraphics[width=1.0\linewidth]{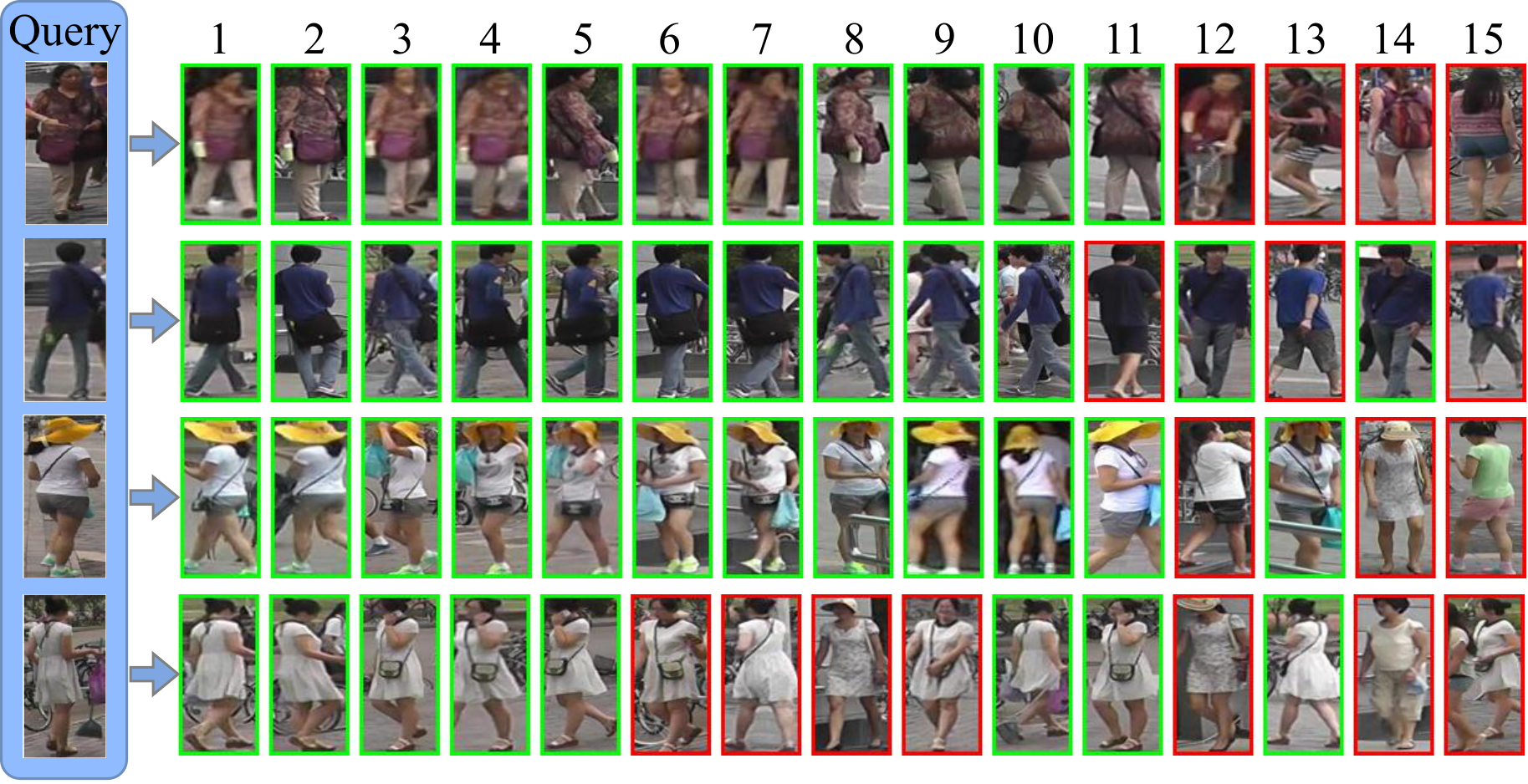}}
\caption{Qualitative Results of our proposed DMF method: top-15 ranking list for some query images on Market-1501 dataset. The images with \textcolor{green}{green} borders belong to the same identity as the given query, and that with \textcolor{red}{red} borders are not matched with query images.}
\label{fig4}
\end{figure*}

\subsection{Ablation study}
\textbf{Multimodal Fusion Training vs. Single-modal Training:}
To prove that multimodal feature fusion strategy effectively improves the performance of the model, we perform ablation study in pre-training stage from both the qualitative and quantitative perspectives.

Firstly, from the quantitative aspect, we select DukeMTMC-reID$\footnote[2]{Note that the DukeMTMC and its derived datasets have been officially removed due to some ethic concerns. Here we include it only for the sake of fine-tuning of real-world datasets. We discourage further usage of DukeMTMC datasets in the future research.}$ as the fine-tuning set, and pre-training with text-only feature, image-only feature and both of them from FineGPR respectively. The detailed results are reported in Table~\ref{table1}. Compared with the baseline model, pre-training with the text-only feature (\textbf{DMF \textit{w/o} Image}) and image-only feature (\textbf{DMF \textit{w/o} Text}) increase the performance by \textbf{+8.6\%} and \textbf{+16.2\%} respectively in terms of mAP accuracy on Market-1501 dataset, which proves both image and text information are the critical part for multimodal training.
When performing pre-training with \textbf{Text \& Image} features together, our method can obtain a remarkable mAP accuracy of 52.6\%, demonstrating the priority of our proposed deep multimodal fusion network.


Secondly, from the qualitative aspect, we also provide a visualization of top-15 ranking results for some given query pedestrian images. As shown in Figure~\ref{fig4}, our DMF method demonstrates the great robustness: regardless of the pose or occupation of these captured pedestrian, deep multimodal features can robustly represent discriminative information of their identities. Specially, the fourth query image is captured in a relatively complicated background. Although the result shows a litter inferiority when compared with first three queries, most of the ranking results are accurate and with high quality. We can obtain her captured images in front view in rank-3, rank-4 and rank-11.
We attribute this surprising result to the effects of multimodal feature fusion,
which establishs relationships when background are complicated or some salient parts are missing.


\begin{figure}[!t]
\centerline{\includegraphics[width=1.0\linewidth]{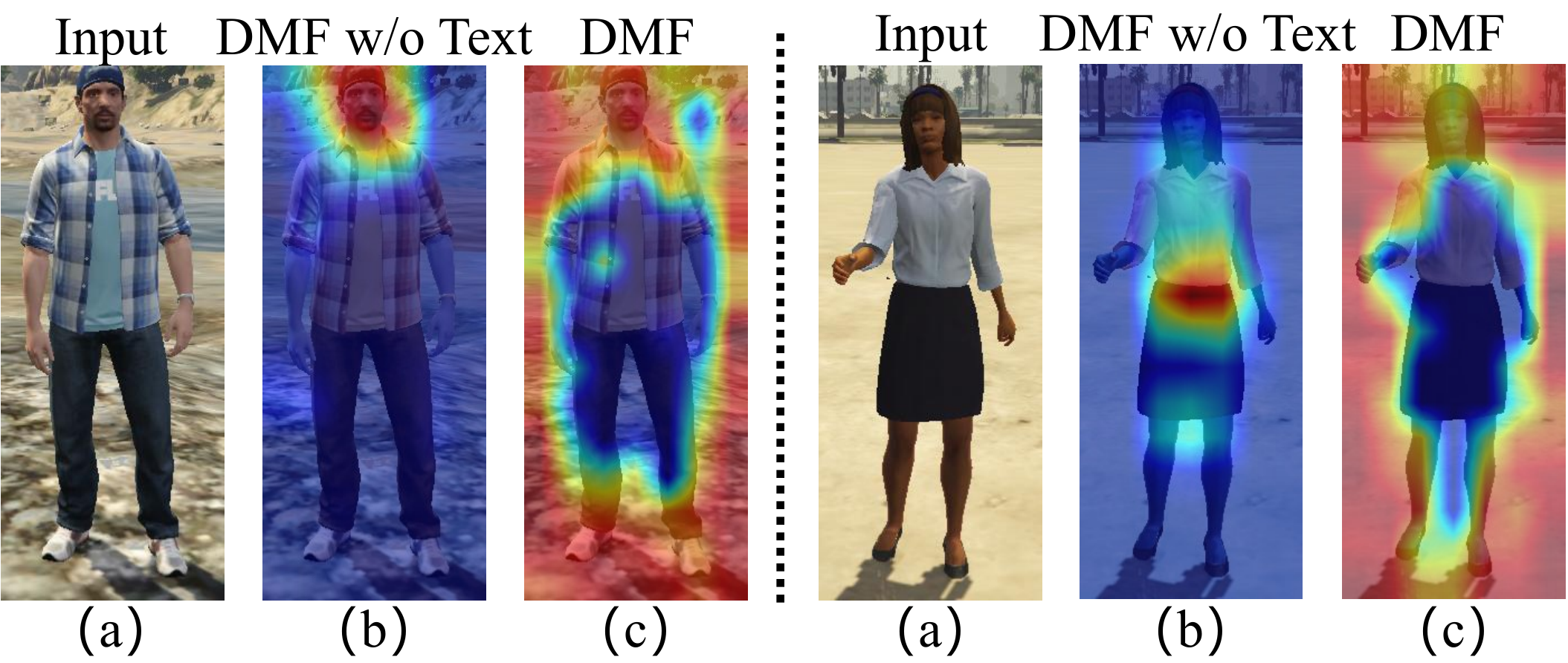}}
\caption{Visualization of feature response between multimodal fusion training and single-modal training. (a) Original input images;
(b) DMF w/o Text (single-modal training); (c) Our DMF method (multimodal training).}
\label{fig5}
\end{figure}

\textbf{Visualization of Feature Response:}
To further explain why multimodal fusion strategy works, we perform in-depth analysis of feature response in DMF method. According to the Figure~\ref{fig5}, which displays some Grad-CAM~\citep{muhammad2020eigen} visualizations of attention maps with pedestrian images between multimodal fusion training and single-modal
training, we can obviously observe that the proposed deep multimodal fusion method mainly attends to relevant global regions or discriminative parts for performing inferences of downstream Re-ID task, while the single-modal method of \textbf{DMF w/o Text} focuses more on the local body parts. In this sense, performing multimodal feature fusion can better take advantage of the complementary properties of Text and Image modalities, which is beneficial to learn a more robust and generalizable model for Re-ID tasks.

\begin{table}[!t]
  \centering
  \caption{Performance comparison with existing methods trained on Market-1501 while tested on CUHK03 and MSMT17 dataset respectively. \textcolor[rgb]{1.00,0.39,0.09}{\textbf{Orange}} indicates the best and \textcolor[rgb]{0.20,0.40,0.80}{\textbf{Blue}} the second best.}
  \small
  \setlength{\tabcolsep}{1.67mm}{
    \begin{tabular}{lcccc}
    \toprule
    \multirow{2}[4]{*}{Methods} & \multicolumn{2}{c}{CUHK03} & \multicolumn{2}{c}{MSMT17} \\
\cmidrule{2-5}          & Rank-1 & mAP   & Rank-1 & mAP \\
    \midrule
    MGN~\citep{wang2018learning, qian2019leader}   & 8.5   & 7.4   & -     & - \\
    MuDeep~\citep{qian2019leader} & 10.3  & 9.1   & -     & - \\
    CBN~\citep{zhuang2020rethinking}   & -     & -     & 25.3  & 9.5  \\
    QAConv~\citep{liao2020interpretable} & 9.9   & 8.6   & 22.6  & 7.0  \\
    QAConv-GS~\citep{liao2022graph} & 16.4  & 15.7  & 41.2  & 15.0  \\
    TransMatcher~\citep{liao2021transmatcher} & \textcolor[rgb]{0.20,0.40,0.80}{\textbf{22.2}}  & \textcolor[rgb]{0.20,0.40,0.80}{\textbf{21.4}}  & \textcolor[rgb]{0.20,0.40,0.80}{\textbf{47.3}}  & \textcolor[rgb]{0.20,0.40,0.80}{\textbf{18.4}} \\
    \midrule
    DMF (Ours)   & \textcolor[rgb]{1.00,0.39,0.09}{\textbf{23.4}}  & \textcolor[rgb]{1.00,0.39,0.09}{\textbf{22.6}}  & \textcolor[rgb]{1.00,0.39,0.09}{\textbf{50.6}}  & \textcolor[rgb]{1.00,0.39,0.09}{\textbf{21.5}} \\
    \bottomrule
    \end{tabular}}%
  \label{table2}%
\end{table}%

\begin{table}[!t]
  \centering
  \caption{Performance comparison with existing methods trained on MSMT17 while tested on Market-1501 and CUHK03 dataset respectively. * represents adopting all (training-set, testing-set and query-set) of MSMT17 dataset for training. \textcolor[rgb]{1.00,0.39,0.09}{\textbf{Orange}} indicates the best and \textcolor[rgb]{0.20,0.40,0.80}{\textbf{Blue}} the second best.}
  \small
  \setlength{\tabcolsep}{1.67mm}{
    \begin{tabular}{lcccc}
    \toprule
    \multirow{2}[4]{*}{Methods} & \multicolumn{2}{c}{Market-1501} & \multicolumn{2}{c}{CUHK03} \\
\cmidrule{2-5}          & Rank-1 & mAP   & Rank-1 & mAP \\
    \midrule
    PCB~\citep{sun2018beyond,yuan2020calibrated}   & 52.7  & 26.7  & -     & - \\
    MGN~\citep{wang2018learning,qian2019leader}   & 48.7  & 25.1  & -     & - \\
    ADIN~\citep{yuan2020calibrated}  & 59.1  & 30.3  & -     & - \\
    SNR~\citep{jin2020style}   & 70.1  & 41.4  & -     & - \\
    CBN~\citep{zhuang2020rethinking}   & 73.7  & 45.0  & -     & - \\
    QAConv-GS~\citep{liao2022graph} & 75.1  & 46.7  & 20.0  & 19.2 \\
    TransMatcher~\citep{liao2021transmatcher} & 80.1  & 52.0  & 23.7  & 22.5 \\
    \multicolumn{1}{l}{DMF (Ours)} & 81.3  & 55.1  & 26.1  & 24.7 \\
    \midrule
    OSNet$^{*}$~\citep{zhou2019omni} & 66.5  & 37.2  & -     & - \\
    QAConv$^{*}$~\citep{liao2020interpretable} & 72.6  & 43.1  & 25.3  & 22.6 \\
    QAConv-GS$^{*}$~\citep{liao2022graph} & 80.6  & 55.6  & 27.2  & 27.1 \\
    TransMatcher$^{*}$~\citep{liao2021transmatcher} & \textcolor[rgb]{0.20,0.40,0.80}{\textbf{82.6}}  & \textcolor[rgb]{0.20,0.40,0.80}{\textbf{58.4}}  & \textcolor[rgb]{0.20,0.40,0.80}{\textbf{31.9}}  & \textcolor[rgb]{0.20,0.40,0.80}{\textbf{30.7}} \\
    \midrule
    DMF$^{*}$ (Ours) & \textcolor[rgb]{1.00,0.39,0.09}{\textbf{82.6}}  & \textcolor[rgb]{1.00,0.39,0.09}{\textbf{58.8}}  & \textcolor[rgb]{1.00,0.39,0.09}{\textbf{34.0}}  & \textcolor[rgb]{1.00,0.39,0.09}{\textbf{32.1}} \\
    \bottomrule
    \end{tabular}}%
  \label{table3}%
\end{table}%

\begin{table}[!t]
  \centering
  \caption{Performance comparison with existing methods trained on RandPerson while tested on Market-1501, CUHK03 and MSMT17 dataset respectively. \textcolor[rgb]{1.00,0.39,0.09}{\textbf{Orange}} indicates the best and \textcolor[rgb]{0.20,0.40,0.80}{\textbf{Blue}} the second best.}
  \small
  \setlength{\tabcolsep}{0.55mm}{
    \begin{tabular}{lcccccc}
    \toprule
    \multirow{2}[4]{*}{Methods} & \multicolumn{2}{c}{Market-1501} & \multicolumn{2}{c}{CUHK03} & \multicolumn{2}{c}{MSMT17} \\
\cmidrule{2-7}          & Rank-1 & mAP   & Rank-1 & mAP   & Rank-1 & mAP \\
    \midrule
    RP Baseline~\citep{wang2020surpassing} & 55.6  & 28.8  & 13.4  & 10.8  & 20.1  & 6.3  \\
    QAConv-GS~\citep{liao2022graph} & 74.0  & 43.8  & 14.8  & 13.4  & 42.4  & 14.4  \\
    TransMatcher~\citep{liao2021transmatcher} & \textcolor[rgb]{0.20,0.40,0.80}{\textbf{77.3}}  & \textcolor[rgb]{0.20,0.40,0.80}{\textbf{49.1}}  & \textcolor[rgb]{0.20,0.40,0.80}{\textbf{17.1}}  & \textcolor[rgb]{0.20,0.40,0.80}{\textbf{16.0}}  & \textcolor[rgb]{0.20,0.40,0.80}{\textbf{48.3}}  & \textcolor[rgb]{0.20,0.40,0.80}{\textbf{17.7}} \\
    \midrule
    DMF (Ours)   & \textcolor[rgb]{1.00,0.39,0.09}{\textbf{78.7}}  & \textcolor[rgb]{1.00,0.39,0.09}{\textbf{52.0}}  & \textcolor[rgb]{1.00,0.39,0.09}{\textbf{21.5}}  & \textcolor[rgb]{1.00,0.39,0.09}{\textbf{19.3}}  & \textcolor[rgb]{1.00,0.39,0.09}{\textbf{52.4}}  & \textcolor[rgb]{1.00,0.39,0.09}{\textbf{18.9}} \\
    \bottomrule
    \end{tabular}}%
  \label{table4}%
\end{table}%

\subsection{Comparison with State-of-the-arts}
In this section, we compare the performance of CNN-based and transformer-based methods in Table~\ref{table2}, Table~\ref{table3} and Table~\ref{table4}. During the experiment, we fine-tune our pre-trained model on Market-1501, MSMT17 and RandPerson dataset, and then evaluate on Market-1501, CUHK03, and MSMT17 respectively. Overall, our method outperforms the state-of-the-art methods such as OSNet~\citep{zhou2019omni}, QAConv~\citep{liao2020interpretable}, QAConv-GS~\citep{liao2022graph} and TransMatcher~\citep{liao2021transmatcher} on several standard benchmark datasets. To be more specific, our DMF method can achieve a rank-1 accuracy and mAP accuracy of 50.6\% and 21.5\% respectively on MSMT17 dataset when fine-tuned on Market-1501 dataset. When compared with the second best approach TransMatcher~\citep{liao2021transmatcher}, our multimodal fusion method DMF leads a significant improvement by \textbf{+3.3\%} and \textbf{+3.1\%} in terms of rank-1 and mAP accuracy respectively on MSMT17 dataset, also with a notable improvement of \textbf{+1.2\%} and \textbf{+1.2\%} on CUHK03 dataset.
Surprisingly, with the synthetic dataset RandPerson for training, our method can obtain a remarkable performance of 78.7\%  and 52.0\% in terms of rank-1 and mAP accuracy on Market-1501 dataset, leading a great improvement of \textbf{+1.4\%}  and \textbf{+2.9\%} when compared with the second best approach TransMatcher~\citep{liao2021transmatcher}. In addition,
as for the cost of training time, our DMF method only costs nearly \textbf{3.5 GPU-days} and \textbf{0.5 GPU-days} for pre-training and fine-tuning respectively even with a large-scale dataset (nearly 2,028,600 image-text pairs) for multimodal fusion, which allows our method more efficient and adaptable in practical scenarios.


\subsection{Discussion}
According to the experiment results, deep multimodal fusion strategy has shown its potential in person Re-ID task. To go even further, we gave an
explanation about two interesting phenomenons observed during the experiment.

Firstly, according to the Table~\ref{table1}, there exists an interesting phenomenon that the performance of \textbf{DMF \textit{w/o} Image} is a litter inferior to the performance of  \textbf{DMF \textit{w/o} Text} (\textit{e.g.} \textcolor[rgb]{0.20,0.40,0.80}{\textbf{38.7\%}} vs. \textcolor[rgb]{1.00,0.39,0.09}{\textbf{46.3\%}} mAP on Market-1501 dataset). We suspect this is due to the image samples contain more high-level and discriminative feature than text words on the backbone of vision transformer network. However, the texture information can further enhance the complementary capacity of image feature in DMF method, which can remarkably improve the performance for generalizable Re-ID system.

Secondly, as shown in Table~\ref{table3}, when tested on Market-1501 dataset, our proposed DMF method can achieve a satisfactory mAP performance of \textbf{55.1\%} and \textbf{58.8\%} when fine-tuned on the MSMT17 and all (training-set, testing-set and query-set) of MSMT17 dataset respectively. Based on this observation, we can conclude that adopting a larger-scale dataset as training samples during the fine-tuning stage
is always beneficial to the system. To this end,
we can drastically improve our performance by enhancing
the scale of training set for pre-training instead of designing more complicated Re-ID model.


\section{Conclusion and Future Work}
\label{sec5}
In this work, we further investigate the possibility of applying Transformers for image retrieval task with multimodal dataset, and then propose a simple but effective multimodal fusion network named
DMF with multimodal datasets to improve the performance of generalizable Re-ID event. To further enhance the robustness of Re-ID model, a multimodal fusion strategy is introduced to translate the data of different modalities into the same feature space in the pre-training stage. During the fine-tuning stage, a realistic dataset is adopted to fine-tune the pre-trained Re-ID model for distribution alignment with real-world. Comprehensive experiments on standard benchmark demonstrate that our method can achieve the state-of-the-art performance in generalizable person Re-ID task. In the future, we will explore the interpretability of this method and apply it to other related computer vision tasks, \textit{e.g.} pose estimation and segmentation.


\bmhead{Acknowledgments}

This work was supported by the National Natural Science Foundation of China under Grant No. 61977045 and No. 81974276.
The authors would like to thank the anonymous reviewers for their valuable suggestions and constructive criticisms.

\section*{Declarations}

\begin{itemize}
\item \textbf{Funding} \\  This work was partially supported by the National Natural Science Foundation of China under Grant No. 61977045 and No. 81974276.
\item \textbf{Conflict of interest} \\  The authors declare that they have no conflict of interest.
\item \textbf{Ethics approval} \\  All procedures performed in studies involving human participants were in accordance with
the ethical standards of the institutional and/or national research committee.
\item \textbf{Consent to participate} \\  All human participants consented for participating in this study.
\item \textbf{Consent for publication} \\  All contents in this paper are consented for publication.
\item \textbf{Availability of data and material} \\  The data used for the experiments in this paper are available online, see Section~\ref{sec4.1} for more details.
\item \textbf{Code availability} \\  The code of this project is publicly available at \url{https://github.com/JeremyXSC/DMF}.
\item \textbf{Authors' contributions} \\  Suncheng Xiang contributed conception and design of the study. Hao Chen contributed to experimental process and evaluated and interpreted model results. Yuzhuo Fu and Dahong Qian obtained funding for the project. Suncheng Xiang, Wei Ran and Zefang Yu reviewed the proof of article. Ting Liu, Dahong Qian and Yuzhuo Fu provided clinical guidance. Suncheng Xiang drafted the manuscript. All authors contributed to manuscript revision, read and approved the submitted version.
\end{itemize}

%
%
%
%
%
%
%

\bibliography{sn-bibliography}


\end{document}